\documentclass[runningheads]{llncs}

\usepackage{graphicx}
\usepackage[T1]{fontenc}
\usepackage[hidelinks]{hyperref}
\usepackage{amsmath}
\usepackage{amssymb}
\usepackage[disable]{todonotes}
\usepackage{booktabs}
\usepackage{cleveref}
\usepackage{enumitem}
\usepackage{makecell}
\usepackage{tikz}

\crefname{figure}{Fig.}{Figures}
\crefname{table}{Table}{Tables}

\begin{document}

\title{Towards Revisiting Visual Place Recognition for Joining Submaps in Multimap SLAM}
\titlerunning{Towards Revisiting VPR for joining Submaps in Multimap SLAM}

\author{Markus Weißflog\inst{1}\orcidID{0009-0003-1163-8755} \and
Stefan Schubert\thanks{The work of Stefan Schubert was supported in part by the German Federal Ministry for Economic Affairs and Climate Action.}\inst{,1}\orcidID{0000-0001-9841-0689} \and
Peter Protzel\inst{1}\orcidID{0000-0002-3870-7429} \and
Peer Neubert\inst{2}\orcidID{0000-0002-7312-9935}%
}

\authorrunning{M. Weißflog et al.}
%
\institute{Process Automation, Chemnitz University of Technology, Chemnitz, Germany \\
\email{markus.weissflog@etit.tu-chemnitz.de} \and
Intelligent Autonomous Systems, University of Koblenz, Koblenz, Germany}

\maketitle              

\begin{tikzpicture}[remember picture,overlay]
    \node[anchor=north, yshift=-1cm] at (current page.north) {
      \parbox{\textwidth}{\scriptsize This preprint has not undergone peer review or any post-submission improvements or corrections. The Version of Record of this contribution is published in \textit{Towards Autonomous Robotic Systems }, and is available online at \url{https://doi.org/10.1007/978-3-031-72059-8_9}\\\rule{\textwidth}{0.4pt}}
    };
\end{tikzpicture}

\begin{abstract}
Visual SLAM is a key technology for many autonomous systems.
However, tracking loss can lead to the creation of disjoint submaps in multimap SLAM systems like ORB-SLAM3.
Because of that, these systems employ submap merging strategies. As we show, these strategies are not always successful.
In this paper, we investigate the impact of using modern VPR approaches for submap merging in visual SLAM.
We argue that classical evaluation metrics are not sufficient to estimate the impact of a modern VPR component on the overall system.
We show that naively replacing the VPR component does not leverage its full potential without requiring substantial interference in the original system.
Because of that, we present a post-processing pipeline along with a set of metrics that allow us to estimate the impact of modern VPR components.
We evaluate our approach on the NCLT and Newer College datasets using ORB-SLAM3 with NetVLAD and HDC-DELF as VPR components. Additionally, we present a simple approach for combining VPR with temporal consistency for map merging.
We show that the map merging performance of ORB-SLAM3 can be improved.
Building on these results, researchers in VPR can assess the potential of their approaches for SLAM systems.

\end{abstract}

\begin{figure}
    \centering
    \includegraphics[width=\textwidth]{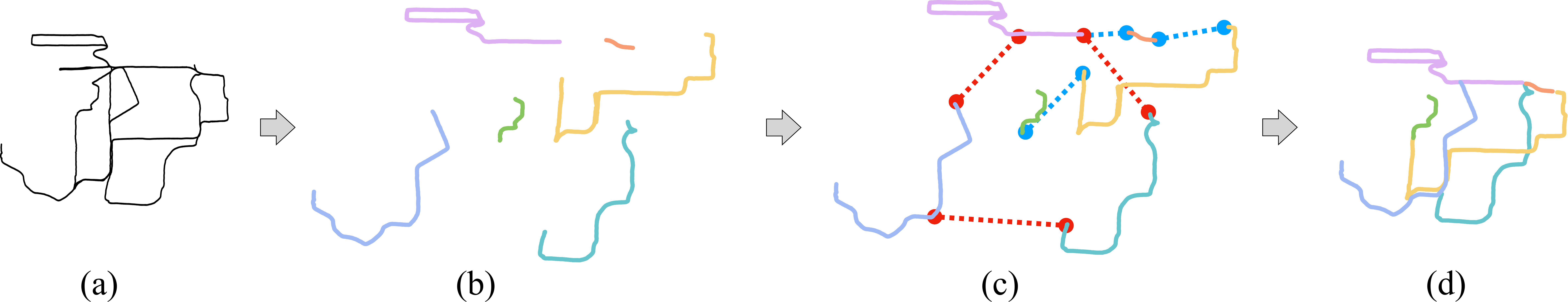}
    \caption{(a) Ground truth trajectory. (b) The output of ORB-SLAM3 is a set of disjoint submaps, with no known transformations between them. (c) The transformations are estimated using VPR (red dotted lines) and temporal consistency (blue dotted lines). (d) The final joined map.}
    \label{fig:introfig}
    \vspace{-0.5cm}
\end{figure}
\section{Introduction}

Visual Simultaneous Localization And Mapping (SLAM) is a key technology for autonomous systems. It has applications in areas like robotics, autonomous driving, and augmented reality. In SLAM, an agent aims to create a map of its environment and tries to localize itself in this map at the same time. The resulting map can be used for surveying, navigation, obstacle avoidance, and other tasks.

However, the agent cannot always accurately estimate its current position. One of the reasons for this is tracking loss, which prevents the agent from estimating its pose in relation to the global map. Tracking loss can occur due to fast motion, textureless regions, occlusions, or various other factors. In multimap visual SLAM systems, the agent has no choice but to create new maps, resulting in a set of disjoint submaps whose relative pose to each other is unknown, as shown in \cref{fig:introfig}. This is a problem for mapping applications, for example, as it is not possible to create a continuous map.

For this reason, modern visual SLAM systems like ORB-SLAM3 \cite{Campos21} use submap merging strategies. As we will show in \cref{sec:orbslam-performance}, these strategies are not always successful, especially on challenging datasets. A promising approach to counteract this problem is Visual Place Recognition (VPR). VPR is the task of recognizing a previously visited place based on images. This can be used to merge submaps. We will show in \cref{sec:vpr_performance} that there are newer VPR approaches that outperform ORB-SLAM3's bag-of-visual-words-based (BOW) approach.

However, as we show in \cref{sec:naive_approach}, by just switching out the VPR component in isolation, ORB-SLAM3 cannot make use of the improved VPR performance. A complete integration into the overall system is required, which would involve considerable implementation effort and changes to the algorithm. Before these changes are made, it would be sensible to estimate the impact of using modern VPR for submap merging on the overall system. This estimation is the goal of our work.
For this, we introduce a simplified experimental setup in \cref{sec:approach}.
We evaluate this setup in \cref{sec:results}.
Finally, we discuss the results in \cref{sec:discussion} and conclude in \cref{sec:conclusion}. We start by discussing the related work in \cref{sec:related_work}. Our main contributions are:
\begin{itemize}
    \item We present a pipeline along with a set of metrics to evaluate the performance of VPR for submap merging in visual SLAM. Our pipeline does not require reparametrization or other modifications to the SLAM system.
    \item We evaluate our approach on two challenging datasets using a state-of-the-art SLAM system with modern VPR approaches.
    \item We present a simple approach for map merging that combines the advantages of modern VPR with temporal consistency.
\end{itemize}

\section{Related Work}
\label{sec:related_work}
\textbf{Visual Simultaneous Localization and Mapping.}\quad
A general introduction to SLAM is given in \cite{Thrun06}.
This paper focuses on monocular visual SLAM, which uses a single camera as the only sensor.
\cite{Sahili23} provides a survey of visual SLAM methods.
Modern SLAM algorithms include maplab 2.0 \cite{Cramariuc23}, VINS \cite{Qin19}, Basalt \cite{Usenko20} and ORB-SLAM3 \cite{Campos21}. \todo{hier mehr?}

\textbf{Visual Place Recognition.}\quad
Yin et al. \cite{Yin22} survey the general problem of place recognition, including VPR.
Schubert et al. \cite{Schubert24} provide a tutorial on VPR where they introduce the general problem, challenges, and evaluation metrics.
Masone and Caputo \cite{Masone21} provide a comprehensive survey on the role of deep learning in VPR.
A central part of VPR are holistic image descriptors. They convert the pixel data of an image into a vector representation, enabling comparison between two images to determine if they show the same location. Holistic descriptors can either be directly computed from the whole image \cite{Arandjelovic16}, or they are an aggregation of local features \cite{Lowry16,Neubert21a}.

Similar to our work, Khaliq et al. \cite{Khaliq23} compare the loop closing component of ORB-SLAM3 with a deep-learning-based VPR method.
VPR-Bench \cite{Zaffar21} benchmarks different VPR metrics.
However, both publications measure performance only on the VPR tasks and do not consider the underlying SLAM system.

    \subsection{ORB-SLAM3}
    We use ORB-SLAM3 \cite{Campos21} for our experiments, as it is considered a highly performant algorithm \cite{Bujanca21,Tourani22}. ORB-SLAM3 is a landmark-based visual-inertial SLAM approach, which can also handle stereo and RGB-D cameras. In essence, ORB-SLAM3 consists of four main threads, which run in parallel:
\begin{enumerate}
    \item The \textit{tracking thread} aims to localize the current frame in the submap using ORB features \cite{Rublee11}. If localization fails, ORB-SLAM3 saves the current submap and starts a new one.
    \item The \textit{local mapping thread} inserts the current keyframe and its new landmarks into the map and optimizes a local window. 
    \item The \textit{loop and map merging thread} searches the database for similar frames to the current keyframe. Upon finding a matching frame (which must satisfy certain checks, see \cref{sec:naive_approach}), a loop or a merge operation is triggered. If the current keyframe and the matching frame are part of the same submap, a loop closure is performed. If they are part of different submaps, the submaps are merged.
    \item In parallel, full bundle adjustment is performed.
\end{enumerate}
\noindent For tracking, ORB-SLAM3 uses ORB features \cite{Rublee11}. These features are also aggregated and used as holistic descriptors in the mapping and map merging thread. The descriptors are computed using the BOW \cite{Sivic2003} implementation DBoW2 \cite{Galvez12}.
In our experiments, we used ORB-SLAM3 in the monocular configuration and only updated the camera parameters according to the datasets.
    \label{sec:orbslam}

\section{Analysis of the Problem}
    \subsection{Loops and Submap Merges in ORB-SLAM3}
    \label{sec:orbslam-performance}
    \begin{figure}[t]
    \includegraphics[width=\textwidth]{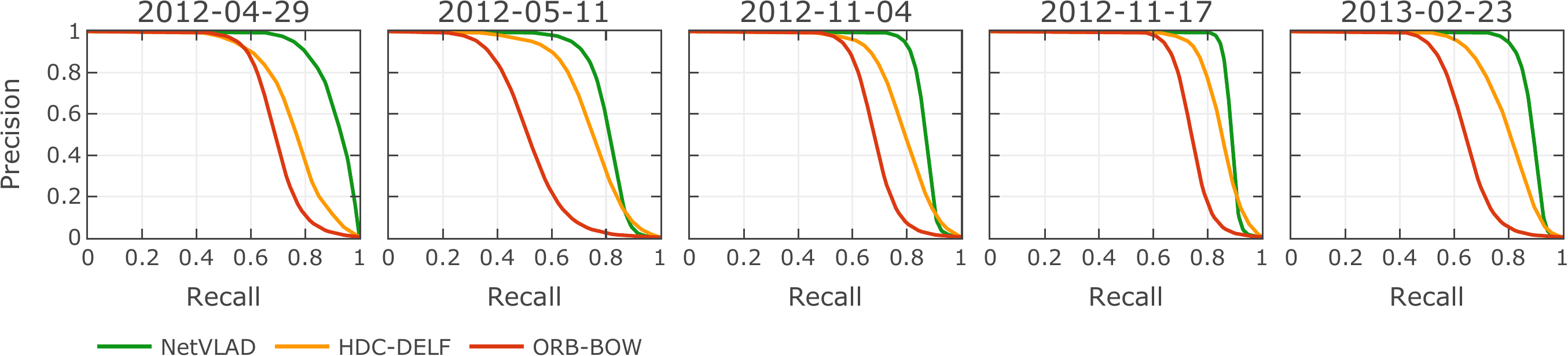}
    \caption{Precision-Recall curves for the analyzed descriptors on the NCLT dataset.  Better performance is indicated by curves that are near the top right corner of the plots.}
    \label{fig:nclt_pr}
\end{figure}
    In this section, we analyze the submap merging performance of ORB-SLAM3 without post-processing.
The results are shown in \cref{tab:orbslam_metrics}. The top part of the table shows the results for the NCLT \cite{CarlevarisBianco16} and Newer College \cite{Ramezani20} datasets, which will be introduced in \cref{sec:datasets}. Noteworthy is the large number of unmerged submaps. The loop and map merging thread is unable to recover the relative transformations between the submaps after tracking loss, which significantly limits mapping capabilities.

We performed the same analysis on the EuRoC \cite{Burri16} dataset, which is shown in the bottom part of the table. This dataset consists of eleven sequences captured using a drone under different conditions and motion patterns. We evaluated all sequences but only show the best (MH\_01) and worst (V2\_02) sequences in \cref{tab:orbslam_metrics}. On this dataset, ORB-SLAM3 has almost no tracking losses, which might be due to the short duration of the sequences.\footnote{The longest sequence lasts for 182 s \cite{Burri16}}
This makes this dataset unsuitable for evaluating submap merging performance. Nevertheless, the number of loops and merges found never surpasses one.
    
    \subsection{Performance of Modern Holistic Descriptors}
    \label{sec:vpr_performance}
    \begin{table}[t]
    \caption{ORB-SLAM3 performance on the analyzed datasets. \textit{Tracked Map} states the proportion of frames that were tracked successfully. \textit{\# Maps} states the number of submaps after the dataset was processed. \textit{\# Loops/ Merges} states the number of successful loop closures or submap merges.}
    \label{tab:orbslam_metrics}
    \setlength\tabcolsep{0pt}
    \footnotesize{
    \begin{tabular*}{\linewidth}{@{\extracolsep{\fill}} llccc}
        \toprule
        \multicolumn{2}{l}{\textbf{Dataset}} & \textbf{Tracked Map}& \textbf{\# Maps} & \textbf{\# Loops/ Merges} \\
        \midrule
        NCLT & 2012-04-29 & 91.7 \% & 23 & 0 \\
        & 2012-05-11 & 97.4 \% & 39 & 0 \\
        & 2012-11-17 & 97.9 \% & 21 & 0 \\
        & 2012-11-04 & 95.9 \% & 24 & 0 \\
        & 2013-02-23 & 99.3 \% & 18 & 2 \\
        \midrule
        \multicolumn{2}{l}{Newer College}  & 80.7 \% & 33 & 1 \\
        \midrule
        EuRoC & MH\_01 & 99.9 \% & 1 & 0 \\
        & \multicolumn{1}{c}{$\vdots$} && \\
             & V2\_03 & 91.5 \% & 2 & 1 \\
        \bottomrule
    \end{tabular*}
    }
    \vspace{-.25cm}
\end{table}
    In this section, we analyze the performance of the BOW-based descriptors used in ORB-SLAM3 in isolation.
We use the widely adapted metrics precision and recall \cite{Schubert24} for evaluation. 
Good performance is indicated by high values for both metrics. Our evaluation setup is similar to \cite{Schubert24}. As can be seen in \cref{fig:nclt_pr}, ORB-BOW is outperformed by the more modern descriptors HDC-DELF and NetVLAD, which will be presented in more detail in \cref{sec:datasets}.
This evaluation hints at the fact that switching out the holistic descriptors in ORB-SLAM3 could lead to better performance in map merging.

However, the performance of the descriptors in isolation does not necessarily translate to better performance in the overall SLAM system. 
For example, matches within a submap contribute to improving the accuracy by providing loop closures; however, they cannot help with joining submaps. Matches between submaps, on the other hand, are crucial for map merging. The difference between these types of matches is not considered by precision and recall. In \cref{sec:approach}, we will present an evaluation metric that takes this difference into account.

    \subsection{Naive Approach: Only Replacing the VPR Descriptor}
    \label{sec:naive_approach}
    \begin{figure}
    \centering
    \includegraphics[width=\textwidth]{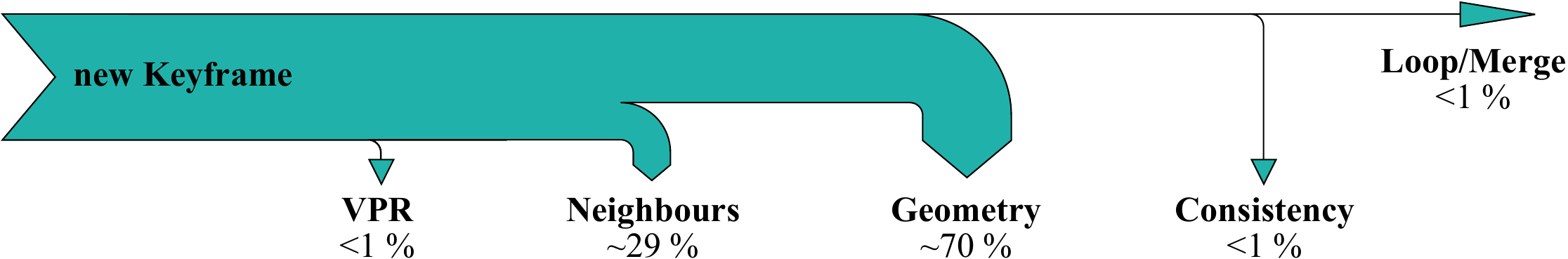}
    \caption{Sankey diagram of the checks performed by ORB-SLAM3. The width of the arrows without the head corresponds to the number of frames that take this path. Downward pointing arrows mean that the loop/ merge is aborted.}
    \label{fig:sankey}
    \vspace{-.25cm}
\end{figure}
    \noindent ORB-SLAM3 performs multiple checks before it performs a loop closure. These checks are visualized in \cref{fig:sankey}. The most important for this work are:
\begin{enumerate}
    \item \textit{VPR}:\label{it:VPR} ORB-BOW descriptors are utilized to find a matching place within the database of keyframes. The check fails if none are found.
    \item \textit{Neighbours}: The current keyframe and the potential match cannot be neighbours in ORB-SLAM3's covisibility graph.
    \item \textit{Geometry}:\label{it:geometry} RANSAC and optimization are used on the ORB features of the current frame and the loop/ merge candidate to estimate the transformation between the two frames. The check is passed if there are enough inlying ORB keypoints.
    \item \textit{Consistency}: The checks have to be passed three times in a row to perform a loop closure/ map merge.
\end{enumerate}

\noindent After a frame passes all checks, the loop closure/ map merge is performed. As can be seen in \cref{fig:sankey}, most loops/ merges are aborted due to the strict check \ref{it:geometry}.
We have experimented with different holistic descriptors for the VPR-check, which has only a marginal effect on \cref{fig:sankey} and no effect on check \ref{it:geometry}.
To benefit from the full potential of modern VPR, the SLAM algorithm would need reparametrization and algorithmic adjustments to its loop/ merge pipeline. HDC-DELF, for example, works on local DELF \cite{Noh17} features, which could also be used during tracking and check \ref{it:geometry}. 
Such deep changes, which could potentially have a significant impact on system performance and involve high implementation efforts, should be tested in a simplified setup before implementation. In the remaining sections, we propose such a setup and evaluate the potential of these changes.

\section{Approach}
\label{sec:approach}

\begin{figure}
    \vspace{-0.25cm}
    \includegraphics[width=\textwidth]{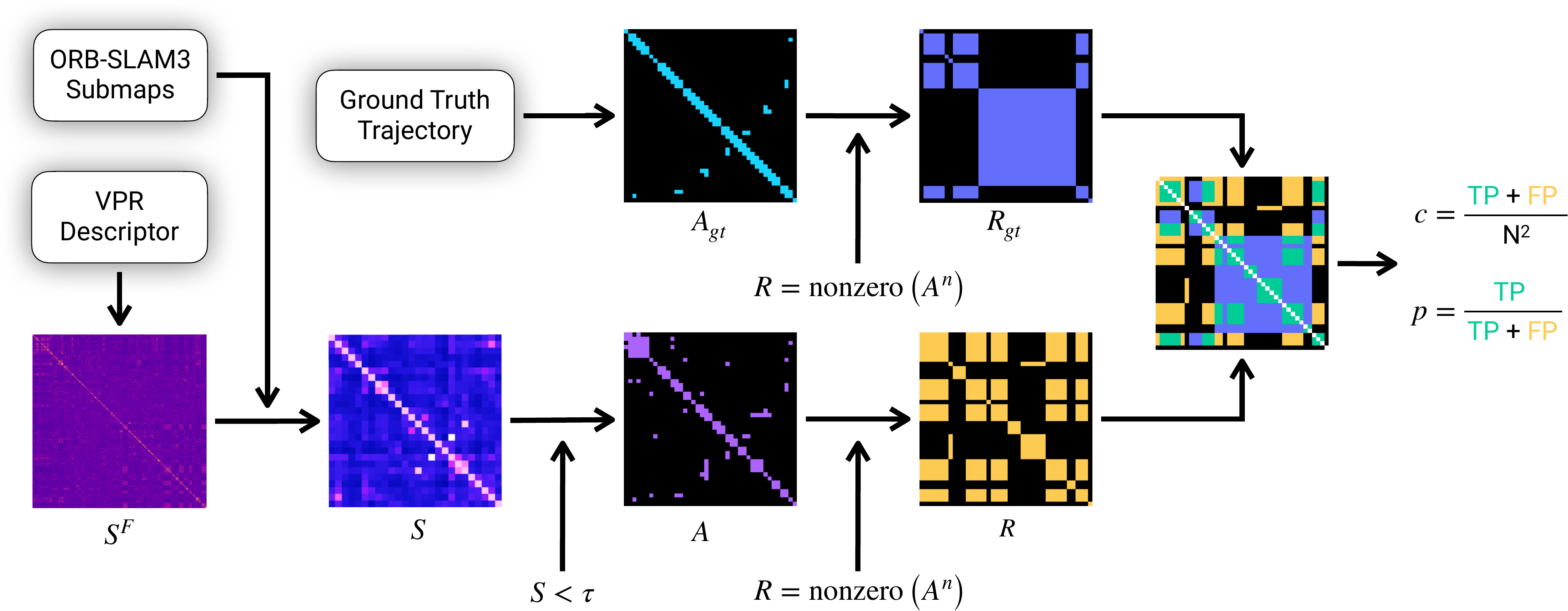}
    \caption{The pipeline shows how precision and coverage are calculated using the example of rule VPR. The VPR descriptors, the ORB-SLAM trajectory, and the ground truth are input to the pipeline. $S^F$ and $S$ are continuous matrices, $A$, $A_{gt}$, $R$, and $R_{gt}$ are boolean. }
    \label{fig:matrices}
    \vspace{-0.25cm}
\end{figure}

\noindent This section describes our evaluation pipeline and the metrics used. An exemplary pipeline is shown in \cref{fig:matrices}. The pipeline starts by predicting the distance between submaps. We use the word distance to refer to the inverse of the similarity. \Cref{sec:setup} describes how VPR and the timestamps of a SLAM trajectory are used to predict submap adjacency matrix $A$. \Cref{sec:metrics} describes how these matrices are used to compute the metrics precision and coverage and how the ground truth is obtained. \Cref{sec:assumptions} summarizes the assumptions we make for our pipeline. Finally, \cref{sec:datasets} describes the descriptors and the datasets used in our experiments.

Our pipeline is a post-processing step for a multimap SLAM system, which outputs a trajectory $\mathcal T$ containing $N$ poses. For this work, we define a trajectory as $\mathcal T = \left\{
            \left(
                t_i, \mathbf p_i, R_i, I_i, j_i
            \right)
        \right\}_i^N
$, where
$t_i$ is the timestamp of pose $i$,
$\mathbf p_i$ and $R_i$ are the position and orientation relative to the submap,
$I_i$ is the camera image,
and $j_i$ is the submap index where frame $i$ is localized, with $j\in \left[1, M\right]$ and $M\ll N$.

    \subsection{Predicting Submap Adjacency}
    \label{sec:setup}
    This section describes how the submap adjacency matrix $A$ is computed, which is required by our post-processing pipeline.
We use the word adjacency to describe that the transformation between two frames or two submaps is known. 
$A$ is a boolean, square, symmetric matrix of size $M\times M$, where $M$ is the number of submaps.
If the transformation between submap $i$ and submap $j$ is known, the submaps are adjacent, and $A$ contains a one at positions $(i, j)$ and $(j, i)$. 
If no post-processing is performed, $A$ is the identity matrix because no transformations between submaps are known.
We analyze three different rules for estimating $A$:

\textbf{Rule Time.}\quad
One natural approach to predict a submap adjacency matrix is to use the timestamps of the submaps. We define the temporal distance between submaps as the difference between the timestamp of the last keyframe of submap $i$ and the timestamp of the first keyframe of submap $j$. By pairwise comparing all timestamps of the trajectory $\mathcal T$, we can create a time distance matrix $S_\mathrm{time}$.

\textbf{Rule VPR.}\quad
Another way to predict the adjacency is to use the visual information of the images $I$. For that, we compute a holistic descriptor for each image using methods from VPR.
We define the visual distance between two images as the distance between their descriptors. Based on this, we can calculate the frame distance matrix $S_{VPR}^F$, which contains at position $(i, j)$ the descriptor distance of the images $I_i$ and $I_j$.
This matrix can be converted into a submap distance matrix $S_\mathrm{VPR}$. At positions $(i, j)$ and $(j, i)$, the matrix $S_\mathrm{VPR}$ contains the smallest visual distance between any two images of submap $i$ and submap $j$. The computation of $S_\mathrm{VPR}$ is visualized in the left part of \cref{fig:matrices}.

\textbf{Combining Time and VPR.}\quad
We test the combination of the two rules using a simple, exemplary algorithm, which can be expanded as required. Using the temporal distance matrix $S_\mathrm{time}$ and the visual distance matrix $S_\mathrm{VPR}$, we define two maps as adjacent, if
\begin{itemize}
    \item their visual distance is below a strict threshold $\tau_{VPR}$, 
    \item or their temporal distance is below a strict threshold $\tau_\mathrm{time}$,
    \item or their temporal distance is below a more relaxed threshold $f_\mathrm{time}\cdot\tau_\mathrm{time}$ and their visual distance is below a relaxed threshold $f_{VPR}\cdot\tau_\mathrm{VPR}$.
\end{itemize}
$f$ is the factor by which the threshold is relaxed. In the later evaluation, we will show exemplary results for the following two combinations of thresholds\footnote{These values were chosen by hand to demonstrate the feasibility of the approach. No parameter optimization was performed}:
\textit{Rule Comb. 1} uses $\tau_\mathrm{time}=2\,s$, $f_\mathrm{time}=10$ and $f_\mathrm{VPR}=2$.
\textit{Rule Comb. 2} uses $\tau_\mathrm{time}=0.5\,s$, $f_\mathrm{time}=10$ and $f_\mathrm{VPR}=4$.
We leave $\tau_\mathrm{VPR}$ as a free parameter.

To convert the distance matrices $S$ into adjacency matrices $A$, a threshold $\tau$ (or $\tau_\mathrm{VPR}$ in the case of the combined rules) is chosen and applied to $S$. All distances below the threshold are set to one, and distances above are set to zero.

\textbf{Ground truth.}\quad
To obtain the ground truth, we define submaps as adjacent if they have two frames whose Euclidean distance (of the ground truth trajectory) and angle of rotation are smaller than two predefined thresholds $\epsilon_\mathrm{dist}=10\,m$ and $\epsilon_\mathrm{rot}=20^\circ$. Similar definitions are typically used for VPR\cite{Zaffar21}.
Thus, the ground truth matrix of adjacent submaps $A_{gt}$ is a boolean matrix of shape ${M\times M}$.

    \subsection{Evaluation Pipeline}
    \label{sec:metrics}
    \textbf{From Adjacency to Reachability.}\quad
The adjacency matrices $A$ can be further processed into reachability matrices $R$. Two submaps are reachable if the transformation between them can be obtained directly or indirectly.\footnote{Indirectly refers to the relationship that, if the transformation between the submaps $(i,j)$ and the transformation between the submaps $(j,k)$ are known, the transformation $(i,k)$ is also known.} 
$R$ is boolean, symmetric, and of size $M\times M$. It is computed as follows:
\begin{align}
    \label{eq:reachability}
    R = \mathrm{nonzero}\left(A^n\right)
\end{align}
The function $\mathrm{nonzero}(\cdot)$ sets all non-zero entries of a matrix to one. $n$ is the number of times the matrix $A$ is multiplied with itself until $R$ converges. This step is visualized in the central portion of \cref{fig:matrices}.

\textbf{Computing Coverage and Precision.}\quad
At first, a weight vector $\mathbf w\in \mathbb N^M$ is extracted from $\mathcal T$, whose elements $w_j$ are defined as the number of frames in submap $j$. A weight matrix $W\in \mathbb N^{M\times M}$ is computed using $W = \mathbf w \mathbf w^T$.
Comparing the ground truth matrix $R_{gt}$ with the predicted reachability matrix $R$ yields the true positive (TP), false positive (FP), true negative (TN), and false negative (FN) matches. From that, coverage and precision can be computed, which is visualized in the right part of \ref{fig:matrices}. \textit{Coverage} describes how much the reachability matrix is filled:
\begin{align}
c = \frac{TP+FP}{N^2}
    = \frac
    {\sum \left( R \odot W \right)}
    {\sum \left(W\right)}
\end{align}
The symbol $\odot$ refers to the Hadamard product, and the symbol $\sum(\cdot)$ refers to the sum over all matrix elements.
\textit{Precision} describes how many of the found matches are correct:
\begin{align}
    p = \frac{TP}{TP+FP}
        = \frac
        {\sum\left( R\odot R_{gt} \odot W\right)}
        {\sum\left( R \odot W\right)}
\end{align}
Note that we exclude the main diagonal of all matrices, as all submaps are always reachable from themselves, and thus precision and coverage would be skewed.
Note also that the adjacency matrix, the estimated reachability matrix, and thus coverage and precision depend all on the threshold value $\tau$ (or $\tau_\mathrm{VPR}$ for the combined rules). By varying this threshold, pairs of precision and coverage values are obtained. All the values together form the precision-coverage curve, which is used for evaluation in the following experiments. The area under the curve (AUC) can be obtained by integration.

    \subsection{Assumptions and Limitations}
    \label{sec:assumptions}
    In this work, we investigate the isolated problem of finding possible matches between submaps of a SLAM system. Because of that, we make several assumptions about the system and the data we work with:
(1) We assume that the SLAM system estimates accurate poses within the submaps.
(2) We assume that the SLAM system can detect tracking loss reliably.
(3) We assume that the transformations between matching frames can be estimated.

    \begin{figure}[t]
    \centering
    \includegraphics[width=\linewidth]{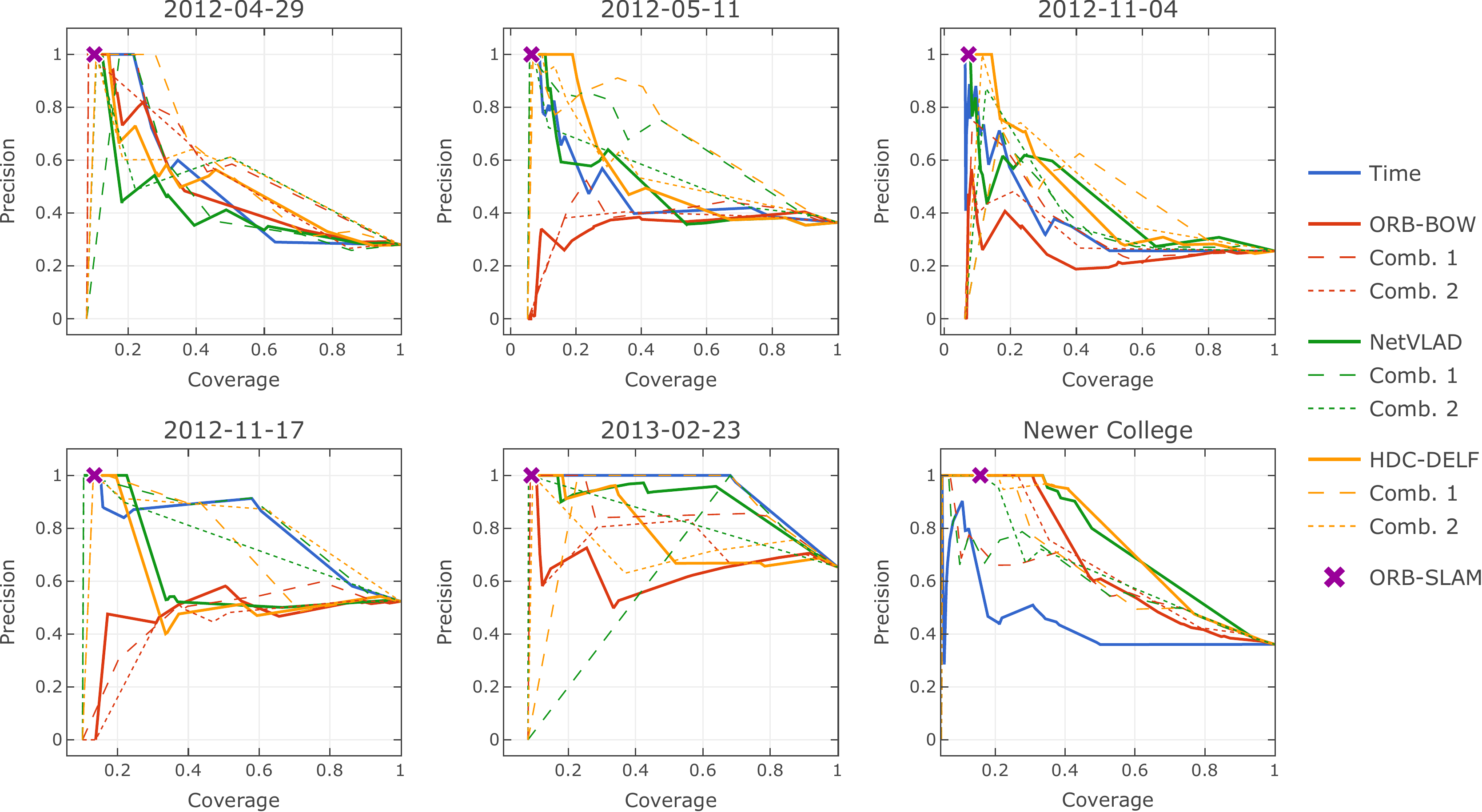}
    \caption{Precision-Coverage curves for the analyzed datasets.}
    \label{fig:precision_coverage}
    \vspace{-0.25cm}
\end{figure}

    \subsection{Experimental Setup}
    \label{sec:datasets}
    \indent \textbf{Datasets.}\quad
We used the NCLT and Newer College datasets.
\textit{NCLT} \cite{CarlevarisBianco16} consists of multiple sequences that were collected onboard a robot covering a large outdoor area of a college campus. We selected five representative sequences that cover different times of the day and different seasons. They have a duration of 40 to 80 minutes.
The \textit{Newer College} dataset \cite{Ramezani20} consists of a single sequence collected using a handheld sensor setup. It covers a large trajectory over a courtyard and a park at New College, Oxford, and has a duration of approximately 44 minutes. We subsample it at 2 Hz.
As we focus on monocular SLAM, we only employ the front camera for NCLT and the left camera for Newer College.
    \par\textbf{Holistic Descriptors.}\quad
We refer to the local BOW-based \cite{Galvez12,Sivic2003} aggregation of ORB features \cite{Rublee11} used by ORB-SLAM3 as \textit{ORB-BOW}.
\textit{HDC-DELF} is a deep learning-based descriptor that uses Hyperdimensional Computing to aggregate DELF features \cite{Noh17} to a holistic descriptor.
\textit{NetVLAD} \cite{Arandjelovic16} is a deep learning-based holistic descriptor as well. In our experiments, we used the implementations and parameters as proposed by the original authors.

\section{Results}
\begin{table}
    \vspace{-.5cm}
    \caption{Area under precision-coverage-curve, with the best-performing rule highlighted in bold, the second-best underlined, and the third-best in italics.}
    \label{tab:auc}
    \centering
    {\footnotesize 
        \begin{tabular*}{\linewidth}{@{\extracolsep{\fill}} lcccccc|cc}
            \toprule
            Rule     & \thead{\footnotesize Newer\\\footnotesize College} & \thead{\footnotesize 2012\\\footnotesize 04-29} & \thead{\footnotesize 2012\\\footnotesize 05-11} & \thead{\footnotesize 2012\\\footnotesize 11-04} & \thead{\footnotesize 2012\\\footnotesize 11-17} & \thead{\footnotesize 2013\\ \footnotesize02-23} & \thead{\footnotesize Mean} & \thead{\footnotesize Worst\\ \footnotesize Case} \\
            \midrule
            Time & 0.418 & 0.461 & 0.454 & 0.334 & \textit{0.709} & \textbf{0.860} & 0.539 & 0.334 \\
            ORB-BOW & 0.637 & 0.453 & 0.337 & 0.240 & 0.425 & 0.594 & 0.448 & 0.240 \\
            Comb. 1 + ORB-BOW & 0.637 & \textit{0.500} & 0.347 & 0.294 & 0.383 & 0.680 & 0.474 & 0.294 \\
            Comb. 2 + ORB-BOW & 0.559 & \textbf{0.552} & 0.370 & 0.330 & 0.426 & 0.785 & 0.504 & 0.330 \\
            NetVLAD & \textbf{0.713} & 0.364 & 0.464 & 0.400 & 0.548 & \underline{0.822} & 0.552 & 0.364 \\
            Comb. 1 + NetVLAD & 0.614 & 0.475 & 0.516 & 0.359 & 0.668 & 0.760 & \textit{0.565} & 0.359 \\
            Comb. 2 + NetVLAD & 0.566 & 0.403 & \underline{0.623} & 0.330 & \textbf{0.733} & 0.559 & 0.536 & 0.330 \\
            HDC-DELF & \underline{0.712} & 0.443 & 0.499 & \underline{0.420} & 0.495 & 0.715 & 0.547 & \underline{0.420} \\
            Comb. 1 + HDC-DELF & \textit{0.706} & 0.491 & \textit{0.528} & \textbf{0.437} & \underline{0.721} & 0.674 & \underline{0.593} & \textbf{0.437} \\
            Comb. 2 + HDC-DELF & 0.620 & \underline{0.526} & \textbf{0.641} & \textit{0.410} & 0.644 & \textit{0.785} & \textbf{0.604} & \textit{0.410} \\
            \bottomrule
        \end{tabular*}
    }
    \vspace{-.2cm}
\end{table}
\label{sec:results}
\noindent Our results can be seen in \cref{fig:precision_coverage} and \cref{tab:auc}.
In \cref{fig:precision_coverage}, the curves are not smooth because a small change in the threshold value $\tau$ could lead to a merge of two large maps, causing coverage and precision to change abruptly by a large amount.\footnote{In all experiments, we iterate over all possible values of $\tau$ to ensure that the smoothest possible curves are obtained.}
On the left of the coverage axis, few highly confident matches are found; on the right, many matches are found, but many of them are false positives.
We see that, except for the Newer College dataset, ORB-BOW is among the lowest-performing rules.
As discussed in \cref{sec:naive_approach}, ORB-SLAM3 employs a very conservative approach to submap merging. This is reflected in the fact that the purple marker in \cref{fig:precision_coverage} has very low coverage but high precision. ORB-SLAM3 is kept as originally proposed in \cite{Campos21}, so there are no parameters that could be varied to get a curve instead of a single point.
The Newer College dataset has large areas of failed tracking (as can be seen by the low tracking rate in \cref{tab:orbslam_metrics}). This causes time-based rules to perform poorer than pure VPR-based rules on this dataset.

\section{Discussion}
\label{sec:discussion}

\Cref{fig:precision_coverage,tab:auc} show a large variation in the results, which could be due to the different characteristics of the datasets.
It can be seen that the ORB-BOW strategy is outperformed in almost all cases.

These results indicate that a modern VPR component offers the potential to improve a SLAM system.
HDC-DELF has a very good mean and worst-case performance, especially in combination with the temporal consistency rule, making it a promising choice for future research.
This VPR approach could also improve other parts of the SLAM pipeline: the local DELF features, for example, could be used for tracking and transformation estimation.
As this work provides an estimation, future research is needed to show the extent to which this potential is realized in a full SLAM pipeline.
We propose that modern VPR methods for map merging as part of a SLAM pipeline should maximize coverage.\todo{diesen Satz vielleicht raus nehmen}

\section{Conclusion}
\label{sec:conclusion}
Multimap SLAM systems like ORB-SLAM3 suffer from tracking loss, which leads to the creation of disjoint submaps without relative pose information between them.
VPR is a potential approach for merging these submaps. However, integrating a modern VPR component into a SLAM system requires considerable modifications to the system.
In this work, we have presented a pipeline to estimate the performance of an improved VPR component for submap merging in visual SLAM. We have evaluated our approach using ORB-SLAM3.
Additionally, we have presented a submap merging approach that combines VPR with temporal consistency for map merging.
Our pipeline does not require reparametrization or changes to the SLAM system.
Our results show that the map merging performance of ORB-SLAM3 can be improved by using modern VPR approaches.
As this work only provides an estimation of possible improvements, future work includes fully integrating the new VPR components into the overall SLAM system to exploit their potential.

\bibliographystyle{splncs04}
\bibliography{references}

\begin{thebibliography}{10}
\providecommand{\url}[1]{\texttt{#1}}
\providecommand{\urlprefix}{URL }
\providecommand{\doi}[1]{https://doi.org/#1}

\bibitem{Arandjelovic16}
Arandjelovic, R., Gronat, P., Torii, A., Pajdla, T., Sivic, J.: {NetVLAD: CNN
  Architecture for Weakly Supervised Place Recognition}. In: Proc. of the Conf.
  on Computer Vision and Pattern Recognition (CVPR).
  \doi{10.1109/TPAMI.2017.2711011}

\bibitem{Bujanca21}
Bujanca, M., Shi, X., Spear, M., Zhao, P., Lennox, B., Luján, M.: {Robust SLAM
  Systems: Are We There Yet?} In: Int. Conf. on Intelligent Robots and Systems
  (IROS). \doi{10.1109/IROS51168.2021.9636814}

\bibitem{Burri16}
Burri, M., Nikolic, J., Gohl, P., Schneider, T., Rehder, J., Omari, S.,
  Achtelik, M.W., Siegwart, R.: {The EuRoC micro aerial vehicle datasets}
  \textbf{35}(10). \doi{10.1177/0278364915620033}

\bibitem{Campos21}
Campos, C., Elvira, R., Rodríguez, J.J.G., M.~Montiel, J.M., D.~Tardós, J.:
  {ORB-SLAM3: An Accurate Open-Source Library for Visual, }visual–inertial,
  and multimap slam  \textbf{37}(6). \doi{10.1109/TRO.2021.3075644}

\bibitem{CarlevarisBianco16}
Carlevaris-Bianco, N., Ushani, A.K., Eustice, R.M.: {University of Michigan
  North Campus long-term vision and lidar dataset}  \textbf{35}(9).
  \doi{10.1177/0278364915614638}

\bibitem{Cramariuc23}
Cramariuc, A., Bernreiter, L., Tschopp, F., Fehr, M., Reijgwart, V., Nieto, J.,
  Siegwart, R., Cadena, C.: {maplab 2.0 – A Modular and Multi-Modal Mapping
  Framework}  \textbf{8}(2). \doi{10.1109/LRA.2022.3227865}

\bibitem{Galvez12}
G\'alvez-L\'opez, D., Tard\'os, J.D.: {Bags of Binary Words for Fast Place
  Recognition in Image Sequences}. IEEE Transactions on Robotics
  \textbf{28}(5) (2012). \doi{10.1109/TRO.2012.2197158}

\bibitem{Khaliq23}
Khaliq, S., Anjum, M.L., Hussain, W., Khattak, M.U., Rasool, M.: {Why ORB-SLAM
  is missing commonly occurring loop closures?} .
  \doi{10.1007/s10514-023-10149-x}

\bibitem{Lowry16}
Lowry, S., Sünderhauf, N., Newman, P., Leonard, J.J., Cox, D., Corke, P.,
  Milford, M.J.: {Visual Place Recognition: A Survey}. IEEE Transactions on
  Robotics  \textbf{32}(1) (2016). \doi{10.1109/TRO.2015.2496823}

\bibitem{Masone21}
Masone, C., Caputo, B.: {A Survey on Deep Visual Place Recognition}
  \textbf{9}. \doi{10.1109/ACCESS.2021.3054937}

\bibitem{Neubert21a}
Neubert, P., Schubert, S., Schlegel, K., Protzel, P.: {Vector Semantic
  Representations as Descriptors for Visual Place Recognition}. In: Proc. of
  Robotics: Science and Systems. \doi{10.15607/RSS.2021.XVII.083}

\bibitem{Noh17}
Noh, H., Araujo, A., Sim, J., Weyand, T., Han, B.: {Large-Scale Image Retrieval
  With Attentive Deep Local Features}. In: Proc. of the Int. Conf. on Computer
  Vision (ICCV)

\bibitem{Qin19}
Qin, T., Pan, J., Cao, S., Shen, S.: {A General Optimization-based Framework
  for Local Odometry Estimation with Multiple Sensors}

\bibitem{Ramezani20}
Ramezani, M., Wang, Y., Camurri, M., Wisth, D., Mattamala, M., Fallon, M.: {The
  Newer College Dataset: Handheld LiDAR, Inertial and Vision with Ground
  Truth}. In: Int. Conf. on Intelligent Robots and Systems (IROS). IEEE.
  \doi{10.1109/iros45743.2020.9340849}

\bibitem{Rublee11}
Rublee, E., Rabaud, V., Konolige, K., Bradski, G.: {ORB: An efficient
  alternative to SIFT or SURF}. In: Int. Conf. on Computer Vision. IEEE.
  \doi{10.1109/iccv.2011.6126544}

\bibitem{Sahili23}
Sahili, A.R., Hassan, S., Sakhrieh, S.M., Mounsef, J., Maalouf, N., Arain, B.,
  Taha, T.: A survey of {Visual SLAM} methods. IEEE Access  \textbf{11} (2023).
  \doi{10.1109/ACCESS.2023.3341489}

\bibitem{Schubert24}
Schubert, S., Neubert, P., Garg, S., Milford, M., Fischer, T.: {Visual Place
  Recognition: A Tutorial} . \doi{10.1109/mra.2023.3310859}

\bibitem{Sivic2003}
{Sivic}, J., {Zisserman}, A.: {Video Google: A text retrieval approach to
  object matching in videos}. In: Int. Conf. on Computer Vision (ICCV) (2003).
  \doi{10.1109/ICCV.2003.1238663}

\bibitem{Thrun06}
Thrun, S., Burgard, W., Fox, D.: {Probabilistic Robotics}. Intelligent robotics
  and autonomous agents, MIT Press, Cambridge, Mass. (2006)

\bibitem{Tourani22}
Tourani, A., Bavle, H., Sanchez-Lopez, J.L., Voos, H.: {Visual SLAM: What Are
  the Current Trends and What to Expect?}  \textbf{22}(23).
  \doi{10.3390/s22239297}

\bibitem{Usenko20}
Usenko, V., Demmel, N., Schubert, D., Stückler, J., Cremers, D.:
  {Visual-Inertial Mapping With Non-Linear Factor Recovery}  \textbf{5}(2).
  \doi{10.1109/LRA.2019.2961227}

\bibitem{Yin22}
Yin, P., Zhao, S., Cisneros, I., Abuduweili, A., Huang, G., Milford, M., Liu,
  C., Choset, H., Scherer, S.: {General Place Recognition Survey: Towards the
  Real-world Autonomy Age}. \doi{10.48550/ARXIV.2209.04497}

\bibitem{Zaffar21}
Zaffar, M., Garg, S., Milford, M., Kooij, J., Flynn, D., McDonald-Maier, K.,
  Ehsan, S.: {VPR-Bench: An Open-Source Visual Place Recognition Evaluation
  Framework with Quantifiable Viewpoint and Appearance Change}. Int. Journal of
  Computer Vision  (2021)

\end{thebibliography}

\end{document}